\documentclass{article}

\usepackage[dblblindworkshop, final]{neurips_2026}
\workshoptitle{BabyVLM: Toward Developmentally Plausible Multimodal Systems}

\usepackage[utf8]{inputenc}
\usepackage[T1]{fontenc}
\usepackage{hyperref}
\usepackage{url}
\usepackage{booktabs}
\usepackage{amsfonts}
\usepackage{nicefrac}
\usepackage{microtype}
\usepackage{xcolor}
\usepackage{multirow}
\usepackage{graphicx}
\usepackage{xspace}
\usepackage{caption}
\usepackage{booktabs}
\usepackage{subcaption}
\usepackage{enumitem}
\usepackage[table]{xcolor}
\usepackage{array}
\usepackage{tcolorbox}
\tcbuselibrary{listings}
\usepackage{amsmath}
\usepackage{makecell}
\usepackage{adjustbox}

\newcolumntype{C}[1]{>{\centering\arraybackslash}p{#1}}
\newcolumntype{L}[1]{>{\raggedright\arraybackslash}p{#1}}
\newcolumntype{R}[1]{>{\raggedleft\arraybackslash}p{#1}}

\definecolor{holisticBlue}{HTML}{C9DAF8}
\definecolor{temporalGreen}{HTML}{D9EAD3}
\definecolor{entityYellow}{HTML}{FFF2CC}
\definecolor{comparativePeach}{HTML}{E4D7F5}

\newcommand{\st}{Spatial Tracking\xspace}
\newcommand{\ta}{Temporal Alignment\xspace}
\newcommand{\cor}{Comparative Reasoning\xspace}
\newcommand{\hs}{Holistic Synthesis}

\newcommand{\stcS}{\colorbox{entityYellow}{Spat. Tracking\xspace}}
\newcommand{\tacS}{\colorbox{temporalGreen}{Temp. Alignment\xspace}}
\newcommand{\corcS}{\colorbox{comparativePeach}{Comp. Reasoning\xspace}}
\newcommand{\hscS}{\colorbox{holisticBlue}{Holi. Synthesis\xspace}}

\newcommand{\newtext}[1]{#1}
\DeclareCaptionFont{appendixblackfont}{\color{black}}

\newcommand{\taskCount}{Object Counting}
\newcommand{\taskRoute}{Route Planning}

\newcommand{\taskOrder}{Sequential Ordering}
\newcommand{\taskSync}{Multi-Angle Synchronization}

\newcommand{\taskReID}{Object Re-identification}
\newcommand{\taskMeasure}{Spatial Measurement}

\newcommand{\taskNumerical}{Numerical Comparison}
\newcommand{\taskKinematic}{Kinematic Comparison}

\newcommand{\taskcCount}{\colorbox{holisticBlue}{Object Counting}}
\newcommand{\taskcRoute}{\colorbox{holisticBlue}{Route Planning}}

\newcommand{\taskcOrder}{\colorbox{temporalGreen}{Sequential Ordering}}
\newcommand{\taskcSync}{\colorbox{temporalGreen}{Multi-Angle Synchronization}}

\newcommand{\taskcReID}{\colorbox{entityYellow}{Object Re-identification}}
\newcommand{\taskcMeasure}{\colorbox{entityYellow}{Spatial Measurement}}

\newcommand{\taskcNumerical}{\colorbox{comparativePeach}{Numerical Comparison}}
\newcommand{\taskcKinematic}{\colorbox{comparativePeach}{Kinematic Comparison}}

\newtcblisting{promptbox}[1][]{
  colback=gray!10,
  colframe=gray!80,
  listing only,
  listing options={
    basicstyle=\ttfamily\small,
    breaklines=true,
    breakatwhitespace=true
  },
  title={#1},
  boxrule=0.5pt,
  arc=2mm,
  width=\linewidth,     %
  left=1mm, right=1mm,  %
  top=1mm, bottom=1mm
}

\colorlet{holisticBlueFrame}{holisticBlue!70!black}
\colorlet{temporalGreenFrame}{temporalGreen!70!black}
\colorlet{entityYellowFrame}{entityYellow!75!black}
\colorlet{comparativePeachFrame}{comparativePeach!70!black}

\tcbset{
  pillar style/.style 2 args={
    colback=#1!45,
    colframe=#2,
    listing only,
    listing options={
      basicstyle=\ttfamily\small,
      breaklines=true,
      breakatwhitespace=true
    },
    boxrule=0.5pt,
    arc=2mm,
    width=\linewidth,
    left=1mm, right=1mm,
    top=1mm, bottom=1mm
  }
}

\newtcblisting{temporalbox}[1][]{pillar style={temporalGreen}{temporalGreenFrame}, title={#1}}

\newtcblisting{spatialbox}[1][]{pillar style={entityYellow}{entityYellowFrame}, title={#1}}

\newtcblisting{comparativebox}[1][]{pillar style={comparativePeach}{comparativePeachFrame}, title={#1}}

\newtcblisting{holisticbox}[1][]{pillar style={holisticBlue}{holisticBlueFrame}, title={#1}}

\title{SYNCR: A Cross-Video Reasoning Benchmark with Synthetic Grounding}

\author{Sara Ghazanfari\thanks{Correspondence: \texttt{sg7457@nyu.edu}}
\hspace{5mm} \textbf{Siddharth Garg}
\hspace{5mm}  \textbf{Prashanth Krishnamurthy}
\hspace{5mm}  \textbf{Farshad Khorrami}
\vspace{0.5cm} \\
New York University
}

\begin{document}

\maketitle
\vspace{-4mm}
\begin{abstract}
Multimodal Large Language Models (MLLMs) have made rapid progress in single-video understanding, yet their ability to reason across multiple independent video streams remains poorly understood. Existing multi-video benchmarks rely largely on human-annotated real-world footage, limiting the precision of spatial, temporal, and physical ground truth and making it difficult to diagnose model failures. We introduce SYNCR, a controlled synthetic benchmark for cross-video reasoning with programmatically verified grounding. Built using Habitat, Kubric, and CLEVRER simulator engines, SYNCR contains 4,000 multi-video question-answer pairs grounded in 4,827 unique videos. It evaluates MLLMs across eight tasks spanning four diagnostic pillars: \textit{Temporal Alignment}, \textit{Spatial Tracking}, \textit{Comparative Reasoning}, and \textit{Holistic Synthesis}.

Our zero-shot evaluation of leading open- and closed-weight MLLMs reveals a substantial gap between current models and humans: the best model achieves only 64.5\% average accuracy, compared to an 89.5\% human baseline. Models perform relatively well on temporal ordering but struggle with precise physical and spatial reasoning, with the best model reaching only 29.8\% accuracy on Kinematic Comparison. We further find that parameter scaling and reasoning-specialized post-training improve temporal alignment capabilities, but do not reliably address fine-grained physical tracking or global spatial synthesis. Finally, a sim-to-real correlation analysis suggests that SYNCR tracks model-level trends on a real-world multi-video benchmark. Data available at \href{https://huggingface.co/datasets/CrossVideoReasoning/SYNCR}{Hugging Face}.
\end{abstract}

\section{Introduction}
\label{sec:intro}

The rapid evolution of Multimodal Large Language Models (MLLMs) has led to substantial progress in image and single-video understanding~\citep{ghazanfari2024emma,wang2025internvl3,bai2025qwen3,ghazanfari2025chain}. Modern architectures can caption dynamic scenes~\citep{li2024llava,xu2024pllava},
answer complex visual queries~\citep{ghazanfari2024towards,fu2025video},
and reason over increasingly long video sequences~\citep{wu2024longvideobench,zhou2024mlvu}. However, human perception and many real-world applications are rarely confined to a single isolated perspective. From multi-angle security feeds~\citep{wu2003multi} to synchronized autonomous-vehicle cameras~\citep{gunn2024lift,li2022bevformer}, intelligent systems must synthesize fragmented spatiotemporal information from multiple independent sources. This motivates a shift from single-video perception to \textit{cross-video reasoning}: the ability to align, compare, and integrate information across disjointed video streams.

Recent efforts have begun to formalize multi-video evaluation through benchmarks such as MVU-Eval~\citep{peng2025mvu} and CVBench~\citep{zhu2025cvbench}. These benchmarks are valuable for measuring broad semantic understanding over real-world footage, but their reliance on human annotation limits their suitability for fine-grained diagnostic evaluation. In particular, human annotators cannot reliably provide exact 3D relative distances, sub-second temporal offsets, camera-aligned object trajectories, or absolute kinematic variables across disjointed views. As a result, model errors can be difficult to attribute: failures may reflect genuine reasoning limitations, ambiguity in the visual evidence, or noise in the ground-truth labels. This annotation bottleneck also limits the scalability of multi-video training data, especially for tasks requiring dense spatial, temporal, or physical supervision.

\begin{figure}
    \centering
    \includegraphics[width=1.0\linewidth,trim={0 3pt 0 28pt},clip]{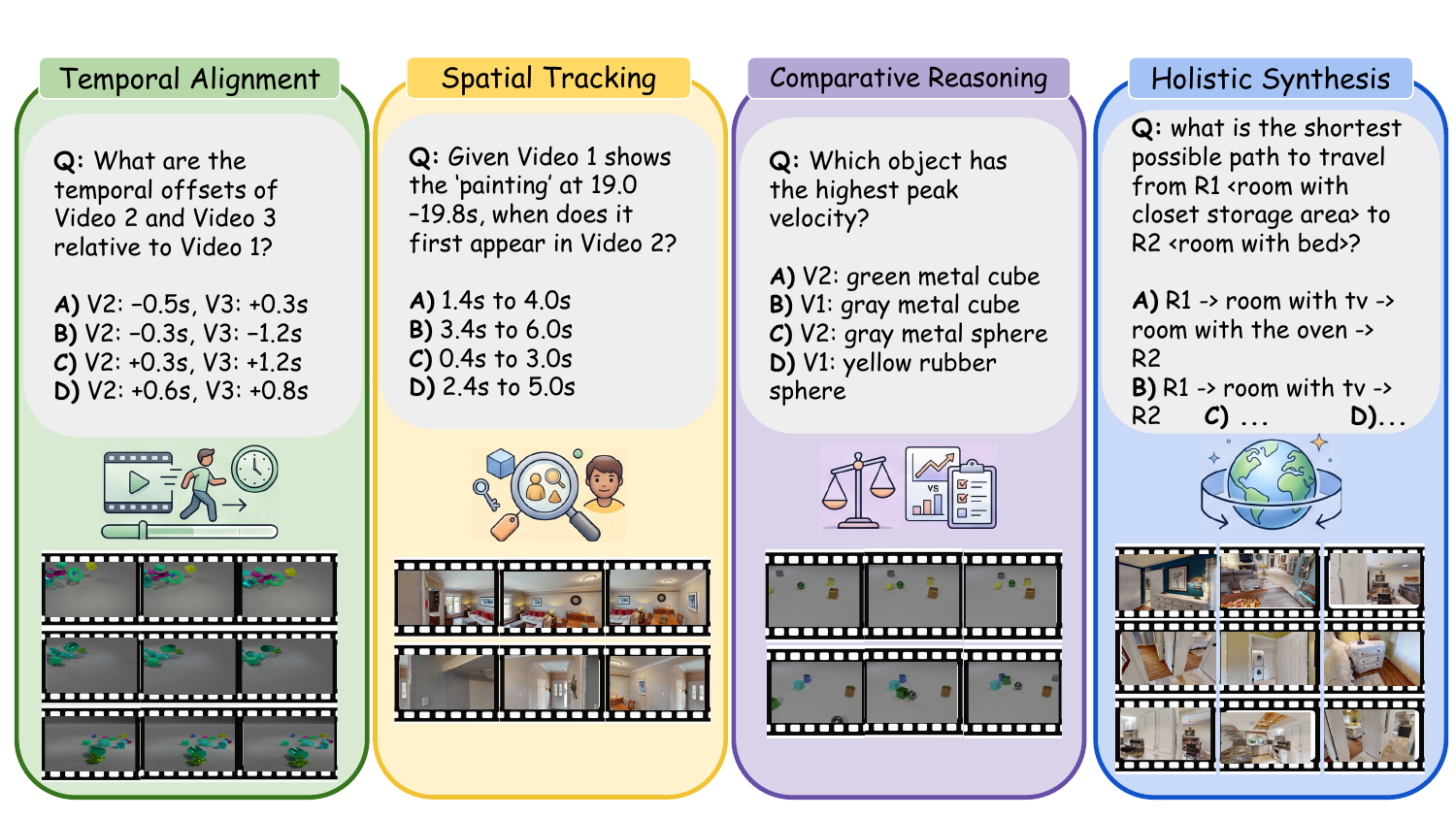}
  \caption{
\textbf{The SYNCR benchmark framework.}
SYNCR evaluates cross-video reasoning through four diagnostic pillars.
\textit{Temporal Alignment} tests synchronization and chronological ordering across unaligned streams.
\textit{Spatial Tracking} evaluates object permanence and cross-view geometry.
\textit{Comparative Reasoning} measures relative physical or numerical properties across videos.
\textit{Holistic Synthesis}
requires integrating fragmented observations into a global scene-level representation.$^1$}

\label{fig:syncr-teaser}
\end{figure}

To address this gap, we develop \textbf{SYNCR} as a controlled simulation-based framework for isolating the core reasoning skills required by multi-video understanding. Rather than using simulation only to render single-view clips, SYNCR generates distinct multi-video subsets by independently leveraging three complementary engines: Habitat~\citep{savva2019habitat}, Kubric~\citep{greff2022kubric}, and CLEVRER~\citep{yi2019clevrer}. By drawing on the specialized simulator state of each individual engine, we construct isolated multi-stream tasks with controlled temporal offsets, camera viewpoints, object identities, physical trajectories, semantic instances, and topological routes.
This design allows us to decompose cross-video understanding into four diagnostic pillars, illustrated in Fig.~\ref{fig:syncr-teaser}: \textit{\ta} for synchronizing and sequencing disjointed events, \textit{\st} for maintaining object permanence and resolving cross-view geometry, \textit{\cor} for comparing kinematic or structural differences across videos, and \textit{\hs} for aggregating fragmented observations into a global scene-level representation. Each pillar is instantiated by two corresponding tasks, yielding eight tasks in total.

We evaluate leading proprietary and open-weight MLLMs, including Gemini-3~\citep{team2023gemini}, GPT-5.4~\citep{singh2025openai}, GPT-6~Astra~\citep{openai2026gpt6astra}, Qwen3-VL~\citep{bai2025qwen3}, and InternVL-3.5~\citep{wang2025internvl3}, in a zero-shot setting. Our results reveal a substantial gap between current models and humans: the top-performing model achieves only 64.5\% average accuracy, compared to an 89.5\% human baseline. While some models perform well on isolated temporal ordering tasks, they struggle with precise physical and spatial synthesis. \taskKinematic~(a \textit{\cor} task) is especially challenging, with no closed-weight model exceeding 20.0\% and the best open-weight model reaching only 29.8\% accuracy. Moreover, parameter scaling and reasoning-specialized post-training enhance temporal alignment, but neither reliably resolves fine-grained physical tracking, cross-video entity reasoning, or global spatial synthesis. Finally, correlations with a real-world multi-video benchmark suggest that SYNCR captures model-level trends beyond simulation.

\vspace{-3mm}
\section{Related Work}
\label{sec:related}

\textbf{Multi-Video Understanding Benchmarks.}
Recent advances in multimodal foundation models have shifted video-language evaluation from isolated single-video perception~\citep{fu2025video,zhou2024mlvu,li2024mvbench} toward more complex multi-video reasoning. Benchmarks such as CVBench~\citep{zhu2025cvbench} and MVU-Eval~\citep{peng2025mvu} introduce benchmarks for evaluating cross-video object association, event correlation, and commonsense reasoning, while CrossVid~\citep{li2026crossvid} scales evaluation of temporal dynamics and comparative analysis with over 5,300 videos and 9,000 QA pairs. These benchmarks provide valuable measures of semantic understanding over real-world footage, but their reliance on human annotation limits both diagnostic precision and scalability. Exact 3D distances, sub-second temporal offsets, camera-aligned trajectories, and continuous kinematic variables are difficult to annotate reliably across disjointed streams, making model errors hard to attribute to either reasoning failures or ground-truth ambiguity.
SYNCR differs from these benchmarks by using simulation not merely as a source of videos, but as a source of structured cross-video supervision.
Finally, while prior datasets evaluate broad multi-video understanding, our framework isolates the temporal, spatial, physical, and topological reasoning skills needed for robust cross-video intelligence.

\textbf{From Single-Video Simulation to Multi-View Generation.}
Synthetic generation has long been used to study physical, causal, and spatial reasoning under controlled conditions. CLEVRER~\citep{yi2019clevrer} uses simulated object interactions to evaluate descriptive, explanatory, predictive, and counterfactual reasoning over videos. Kubric~\citep{greff2022kubric} provides a scalable simulation and rendering pipeline for generating photorealistic videos with pixel-level and 3D ground-truth annotations, including segmentation, optical flow, and object trajectories. Embodied AI platforms such as Habitat~\citep{savva2019habitat,szot2021habitat,puig2023habitat} enable controlled rendering and navigation in complex 3D environments, supporting tasks that require spatial and topological understanding. While these platforms provide precise simulator-derived ground truth, they have primarily been used to construct single-video or single-perspective evaluations. SYNCR extends this line of work by leveraging CLEVRER, Kubric, and Habitat into a unified benchmark for synchronized, multi-camera, and temporally overlapping video streams. This enables diagnostic evaluation of cross-video reasoning with simulator-derived grounding.

\section{SYNCR Framework}
\label{sec:framework}
Complex multi-video understanding requires moving from local spatiotemporal grounding to higher-level comparison and scene synthesis. We formalize this structure through four diagnostic pillars inspired by developmental psychology and prior work on physical, relational, and world-model reasoning~\citep{spelke2007core,kirchhoff2019extended,greff2020binding,yi2019clevrer,santoro2017simple,ha2018world,lecun2022path}. At the foundation, models must establish spatiotemporal anchors through \st and \ta, maintaining object identity and temporal correspondence across disjointed streams. These anchors support \cor, where models compare parallel events by tracking kinematic, structural, or numerical differences across videos, and \hs, where models aggregate fragmented and partially overlapping observations into a unified scene-level representation.

\footnotetext[1]{The pillar icons in this figure were generated using Google's Gemini 3 Flash Image model for illustration only.}

\textbf{\ta.}
Models must establish a shared chronological frame of reference across independent video streams. This pillar evaluates whether a model can align unsynchronized views and sequence fragmented clips into a coherent timeline. We instantiate it with two tasks. {\taskSync} requires deducing temporal offsets between unaligned videos of the same event captured from different perspectives.
{\taskOrder} requires reconstructing the chronological order of shuffled clips by tracking object motion and kinematic continuity.

\textbf{\st.} Models must maintain object permanence as entities move across disjointed visual contexts, despite changes in viewpoint, lighting, and camera extrinsics. This goes beyond 2D appearance matching by requiring object identity to be resolved through cross-view geometry. We instantiate this with two tasks. {\taskReID} asks the model to track a target from a reference video and identify its temporal window in a second stream. {\taskMeasure} requires estimating relative 3D proximity in dynamic multi-camera scenes using cross-view object anchors.

\textbf{\cor.}
This pillar evaluates whether models can analyze parallel events across videos to identify structural differences, kinematic patterns, or numerical discrepancies. Building on temporal and spatial grounding, it requires tracking multiple evolving physical states and judging their relative properties. We instantiate it with two tasks. {\taskKinematic} requires comparing object trajectories across similar interactions to identify the object with the maximum velocity. {\taskNumerical} requires aggregating discrete events, such as collisions, across videos and computing the numerical difference between them.

\textbf{\hs.}
This pillar evaluates whether models can combine partially overlapping observations into a coherent global representation while resolving redundancy and missing visual context. We instantiate it with two tasks. {\taskCount} requires counting distinct entities across overlapping trajectories while avoiding double-counting repeated observations. {\taskRoute} requires constructing an implicit topological map from partial observations to identify the shortest navigable path between disconnected regions.

\vspace{-2mm}
\section{Dataset Generation with Synthetic Grounding}
\label{sec:data}

The diagnostic evaluation of the cognitive pillars outlined in Sec.~\ref{sec:framework} requires ground truth that is precise, consistent, and available across temporal, spatial, physical, and topological variables. SYNCR obtains this grounding by combining three complementary simulation environments: Habitat for semantic navigation and topological synthesis, Kubric for multi-camera spatial and temporal reasoning, and CLEVRER for kinematic and collision-based reasoning.  Our framework applies a three-step generation pipeline across all engines: (1) simulator state extraction to gather precise, frame-aware variables; (2) synchronized video rendering and sequence cropping; and (3) programmatic QA formulation. This ensures that the benchmark purely evaluates a model's capacity for efficient visual alignment and cross-video reasoning, completely isolated from human annotation errors.

\vspace{-3mm}
\subsection{Habitat: Semantic Navigation and Topological Synthesis}

Habitat~\citep{savva2019habitat} with HM3D~\citep{ramakrishnan2021habitat} provides RGB egocentric observations, semantic instance annotations, and navigation meshes for large indoor environments. We use the 216 HM3D scenes with dense semantic annotations, illustrated in Fig.~\ref{fig:habitat-semantic} of App.~\ref{app:dataset_details}. These annotations allow SYNCR to ground object visibility, semantic instance identity, and navigable connectivity directly in simulator state.

\textbf{Video Generation.}
We generate continuous egocentric trajectories over the Habitat NavMesh and smooth them using cubic spline interpolation~\citep{virtanen2020scipy}. Each rendered sequence contains 100 frames at 5 FPS and ends with a panoramic sweep to provide additional endpoint context. We use two trajectory-generation strategies. First, geodesic sampling selects spatially separated trajectories to expose distinct regions and object instances across videos, supporting object re-identification and object counting. Second, topological graph routing samples connected paths over the NavMesh and slices them into separate clips while preserving endpoint continuity, supporting route planning.

\textbf{\taskcReID.}
This task takes two input videos, designated as the reference and the target. Ground truth is extracted programmatically from frame-wise object ID matrices to track the exact timestamps a target instance ID appears in the reference video, logging its first appearance window in the target video. To eliminate ambiguity from transient pixels or structural background elements, we retain an object only if its continuous appearance lasts for at least five frames (see Fig.~\ref{fig:prompt_habitat_reid} for the QA template).

\textbf{\taskcCount.}
This task requires the model to aggregate observations across three input videos. Ground-truth counts are derived by extracting the exact number of unique semantic instance IDs for two or three queried object categories across all generated trajectories, with each category capped at 10 instances. To reduce visual ambiguity, we enforce a strict pixel-occupancy threshold, requiring the target instance to occupy at least 5\% of the total pixels in at least one frame. Finally, the dataset is dynamically rebalanced to prevent a statistical dominance of a specific count (see Fig.~\ref{fig:prompt_habitat_count} for the QA template).

\textbf{\taskcRoute.}
This task requires the model to infer navigable connectivity from three input videos. Ground-truth shortest path between a starting and ending region is extracted from NavMesh connectivity logs, requiring a minimum path length of three nodes. Because regions in the simulator are defined
by numerical IDs, we map these nodes to natural language descriptions.
Specifically, each room is identified by referencing three of its objects that have the lowest total frequency across the entire scene. Moreover, common background elements and structural noise words are excluded from these sets to ensure distinctive visual routing cues (see Fig.~\ref{fig:prompt_habitat_route} for the QA template).
\subsection{Kubric: Multi-Camera Temporal and Spatial Grounding}

Kubric~\citep{greff2022kubric} provides controllable rendering of dynamic object interactions with access to camera parameters, object visibility, and 3D object state. While Habitat supports semantic navigation
Kubric fills the setting of controlled multi-camera object motion with known viewpoints and 3D positions.

\textbf{Video Generation.}
We simulate dynamic interactions among 10--15 KuBasic geometric primitives using PyBullet. Objects are initialized in a $10 \times 10$ region with randomized velocities and a frictionless floor to encourage sustained motion across 60-frame sequences at 12 FPS. Scenes are rendered with Blender's Cycles engine at $256 \times 256$ resolution. Each simulation is captured by a synchronized three-camera rig with distinct viewpoints, reducing field-of-view overlap while preserving a shared underlying 3D scene. This setup provides known crop offsets, camera viewpoints, object visibility, and 3D coordinates for task construction.

\textbf{\taskcSync.}
This task requires a model to infer temporal offsets between three asynchronous videos of the same physical event. We extract randomized 3-second crops from 5-second master simulations and ask the model to determine the starting timestamps of two target videos relative to a reference video. Independent crop sampling allows target videos to start before or after the reference, producing both negative and positive offsets (see Fig.~\ref{fig:prompt_kubric_sync} for the QA template).

\textbf{\taskcMeasure.}
This task requires a model to identify physical proximity in 3D space given two videos that capture the same 3D event from different camera locations. Given a target object and a temporal anchor, the model must select the object closest to the target in simulator coordinates. The temporal anchor is defined by a visible event: the frame in which a designated event object exits one camera view. We filter target and candidate objects using pixel-occupancy thresholds, then compute Euclidean distances from simulator-derived 3D object coordinates (see Fig.~\ref{fig:prompt_kubric_spatial_measurement} for the QA template).

\vspace{-2mm}
\subsection{CLEVRER: Temporal, Kinematic, and Physical Reasoning}

CLEVRER~\citep{yi2019clevrer} provides simulated multi-object collision videos with frame-level annotations of object identity, attributes, trajectories, velocities, and collision events. Unlike Habitat and Kubric, we do not render new videos; instead, we reuse CLEVRER's existing video corpus and derive cross-video tasks from its simulator annotations. This allows SYNCR to evaluate temporal sequencing, continuous motion comparison, and discrete event counting using structured physical state.

\textbf{\taskcOrder.}
This task requires the model to reconstruct a continuous physical event from shuffled video fragments, generated by splitting each 128-frame simulation into 4 equal-length clips separated by random gaps of 0.3--0.6 seconds, so that no two clips share a boundary frame. The correct answer is the chronological order of the clips. Every ordering is equally frequent as an answer, and distractor options are sampled in proportion to answer frequencies (see Fig.~\ref{fig:prompt_clevrer_order} for the QA template).

\textbf{\taskcKinematic.}
This task evaluates whether a model can compare continuous object motion across videos. Given two  videos, the model must identify the object with the highest peak velocity. We compute maximum instantaneous velocities from CLEVRER's frame-level trajectories and retain examples only when the leading object exceeds alternatives by a minimum margin, reducing ambiguity from objects with visually similar speeds (see Fig.~\ref{fig:prompt_clevrer_kinematic} for the QA template).

\textbf{\taskcNumerical.}
This task takes two input videos and the model must identify which video contains the most collisions and compute the numerical difference between the highest and second-highest collision counts. Ground-truth counts are derived directly from CLEVRER's collision-event annotations, explicitly accounting for tie cases, which make up one fifth of the items; each question has five answer options (see Fig.~\ref{fig:prompt_clevrer_numerical} for the QA template).

\subsection{Dataset Statistics} Table~\ref{tab:dataset_stats} details the composition of the SYNCR benchmark. In total, the benchmark comprises 4,000 question-answer pairs (500 per task) grounded in 4,827 unique videos generated across three simulation engines. Each example requires reasoning over two to four independent videos. Distributed across four diagnostic pillars and eight distinct tasks, SYNCR goes beyond narrow synthetic evaluations to provide a comprehensive, multi-engine testbed for cross-video reasoning at scale.

\begin{table}[!htbp]
    \centering
    \caption{\textbf{SYNCR dataset statistics.} Unique videos count distinct video files; Videos/Inp.\ is the number of videos per question; Clip Len.\ is the length of each clip shown to the model. Per-task counts are the distinct videos used by that task; the total is the union over tasks, which is smaller than their sum (5,688) because some videos are shared between tasks.}
    \label{tab:dataset_stats}
    \adjustbox{max width=\textwidth}{%
    \begin{tabular}{ll|c|cc|cc}
        \toprule
        \textbf{Category} & \textbf{Task} & \textbf{Simulation Engine} & \textbf{Unique Videos} & \textbf{QA Pairs} & \textbf{Videos/Inp.} & \textbf{Clip Len. (s)} \\
        \midrule
        \rowcolor{temporalGreen} & Sequential Ordering & CLEVRER & 500 & 500 & 4 & 0.87 \\
        \rowcolor{temporalGreen} \multirow{-2}{2.5cm}{\ta} & Multi-Angle Synchronization & Kubric & 1,500 & 500 & 3 & 3.0 \\
        \midrule
        \rowcolor{entityYellow} & Object Re-identification & Habitat & 72 & 500 & 2 & 20.0 \\
        \rowcolor{entityYellow} \multirow{-2}{2.5cm}{\st} & Spatial Measurement & Kubric & 558 & 500 & 2 & 5.0 \\
        \midrule
        \rowcolor{comparativePeach} & Numerical Comparison & CLEVRER & 1,000 & 500 & 2 & 5.12 \\
        \rowcolor{comparativePeach} \multirow{-2}{2.5cm}{\cor} & Kinematic Comparison & CLEVRER & 837 & 500 & 2 & 5.12 \\
        \midrule
        \rowcolor{holisticBlue} & Object Counting & Habitat & 354 & 500 & 3 & 20.0 \\
        \rowcolor{holisticBlue} \multirow{-2}{2.5cm}{\hs} & Route Planning & Habitat & 867 & 500 & 3 & 20.0 \\
        \midrule
        \multicolumn{2}{l}{\textbf{Total / Average}} & \textbf{-} & \textbf{4,827} & \textbf{4,000} & \textbf{-} & \textbf{9.89} \\
        \bottomrule
    \end{tabular}
    }
\end{table}

\section{Experiments}
\label{sec:experiments}

We design our experiments to test whether current MLLMs can form coherent representations across multiple disjointed video streams, rather than merely solve isolated single-video perception problems. SYNCR enables this analysis because each task is tied to programmatically verified temporal, spatial, physical, or topological ground truth. Our evaluation therefore addresses four questions: how large the gap is between current MLLMs and human performance; which cross-video reasoning skills are most fragile; whether parameter scaling and reasoning-specialized post-training alleviate these failures; and whether synthetic task performance aligns with trends on real-world multi-video benchmarks.

\subsection{Experimental Setup}
\label{sec:experimental_setup}

\textbf{Evaluated Models.}
We evaluate a diverse set of recent open-weight and closed-weight MLLMs. Our open-weight suite includes 17 checkpoints from Qwen2.5-VL~\citep{Qwen2.5-VL}, Qwen3-VL~\citep{bai2025qwen3}, InternVL3~\citep{zhu2025internvl3}, InternVL-3.5~\citep{wang2025internvl3}, and LLaVA-OneVision~\citep{li2024llava}, spanning 0.5B to 72B parameters, together with the reasoning-specialized Qwen3-VL-8B-Thinking.
We also evaluate four proprietary systems, Gemini-3-Flash~\citep{team2023gemini}, Gemini-3.1-Pro~\citep{gemini31pro2026}, GPT-5.4~\citep{singh2025openai}, and GPT-6~Astra~\citep{openai2026gpt6astra}, as strong closed-weight baselines for current multimodal reasoning capabilities.
Together, this model suite allows us to compare performance across architecture families, scale regimes, and open- versus closed-weight systems.

\textbf{Human Baseline.}
To estimate a human reference point, we use the first 100 question-answer pairs from each of the eight SYNCR tasks, the same items given to the proprietary models, yielding 800 examples in total. The sample is balanced across tasks so that each cognitive pillar contributes equally to the final estimate. Four graduate students independently answer each question using the same multiple-choice format given to the models.
Finally, we compute the final human accuracy using majority vote across annotators.

\textbf{Evaluation Protocol.}
All models are evaluated in a zero-shot multiple-choice setting. For each example, we provide the model with the corresponding set of input videos, explicitly labeled in order as \textit{Video 1}, \textit{Video 2}, etc., followed by the task question and answer choices. This labeling is kept consistent across all models to preserve the intended cross-video references in the question. We use the same answer-choice format for all models and extract the predicted option with a shared answer parser (App.~\ref{app:evaluation_details}). All models are evaluated with deterministic decoding by setting temperature to 0.

\begin{table*}[!t]
     \centering
    \caption{\textbf{Zero-shot accuracy (\%) on SYNCR.} Avg is the unweighted mean across eight tasks. The 17 standard open-weight checkpoints and the Thinking checkpoint use 500 questions per task; proprietary models$^\dagger$ and the four-person human majority vote use the same first 100. Bold and underline mark the best and second-best results among the 17 standard open-weight checkpoints. Chance is 25\%, except for \textsc{Num} (20\%).}

  \label{tab:main_results}
  \resizebox{1.0\textwidth}{!}{
    \begin{tabular}{L{34mm}*{8}{C{10mm}}C{6mm}}
    \toprule
    \multirow{2}{*}{\textbf{Model}} & \multicolumn{2}{c}{\textbf{\tacS}} & \multicolumn{2}{c}{\textbf{\stcS}} & \multicolumn{2}{c}{\textbf{\corcS}} & \multicolumn{2}{c}{\textbf{\hscS}} & \multirow{2}{*}{\textbf{Avg}} \\
    \cmidrule(lr){2-3} \cmidrule(lr){4-5} \cmidrule(lr){6-7} \cmidrule(lr){8-9}
    & {\footnotesize\textsc{Sync}} & {\footnotesize\textsc{Order}} & {\footnotesize\textsc{ReID}} & {\footnotesize\textsc{Meas}}  & {\footnotesize\textsc{Num}} & {\footnotesize\textsc{Kin}} & {\footnotesize\textsc{Count}} & {\footnotesize\textsc{Route}} & \\
    \midrule
Human & 100.0 & 100.0 & 92.0 & 76.0 & 100.0 & 68.0 & 92.0 & 88.0 & 89.5 \\

Gemini-3-Flash$^\dagger$ & 28.0 & 84.0 & 72.0 & 32.0 & 24.0 & 4.0 & 28.0 & 48.0 & 40.0 \\
Gemini-3.1-Pro$^\dagger$ & 60.0 & 96.0 & 68.0 & 48.0 & 16.0 & 8.0 & 32.0 & 40.0 & 46.0 \\
GPT-5.4$^\dagger$ & 36.0 & 64.0 & 48.0 & 36.0 & 20.0 & 16.0 & 32.0 & 44.0 & 37.0 \\
GPT-6 Astra$^\dagger$ & 88.0 & 100.0 & 88.0 & 52.0 & 60.0 & 20.0 & 48.0 & 60.0 & 64.5 \\
    \midrule
    \multicolumn{10}{l}{\textbf{Model Size $\leq$ 4B}} \\
    \midrule
LLaVA-OV 0.5B          & 25.8 & 24.4 & 26.4 & 23.0 & 16.0 & 24.0 & 25.4 & 25.4 & 23.8 \\
InternVL3.5 1B         & 24.4 & 25.8 & 24.6 & 24.4 & 20.0 & 27.6 & 22.8 & 27.4 & 24.6 \\
InternVL3.5 2B         & 26.8 & 25.2 & 26.0 & 31.2 & 26.8 & \underline{29.6} & 28.4 & 28.0 & 27.8 \\
Qwen3-VL 2B            & 24.2 & 28.4 & 33.6 & 29.4 & 21.4 & 20.2 & 25.6 & 23.2 & 25.8 \\
InternVL3.5 4B         & 26.4 & 30.6 & 26.2 & 24.8 & 29.2 & 24.2 & 23.6 & 25.6 & 26.3 \\
Qwen3-VL 4B            & 26.2 & 49.4 & 46.4 & \underline{38.0} & 26.6 & 20.0 & 28.8 & 28.0 & 32.9 \\
    \midrule
    \multicolumn{10}{l}{\textbf{4B $<$ Model Size $\leq$ 8B}} \\
    \midrule
LLaVA-OV 7B            & 27.2 & 21.6 & 26.6 & 27.6 & 20.8 & 24.2 & 26.0 & 23.6 & 24.7 \\
Qwen2.5-VL 7B          & 25.0 & 36.8 & 25.8 & 29.8 & 22.4 & 21.2 & 27.2 & 25.6 & 26.7 \\
InternVL3 8B           & 26.0 & 35.4 & 25.0 & 29.6 & 26.6 & 26.0 & 27.2 & 20.0 & 27.0 \\
InternVL3.5 8B         & 26.4 & 39.4 & 26.0 & 31.4 & 31.6 & \textbf{29.8} & 26.4 & 24.0 & 29.4 \\
Qwen3-VL 8B            & 29.6 & 56.4 & \underline{47.6} & 32.6 & 29.6 & 17.2 & \underline{30.2} & 29.4 & 34.1 \\
Qwen3-VL 8B Thinking & 23.8 & 76.6 & 47.0 & 24.0 & 23.0 & 20.6 & 28.6 & 32.0 & 34.4 \\
    \midrule
    \multicolumn{10}{l}{\textbf{Model Size $>$ 8B}} \\
    \midrule
InternVL3.5 14B        & \textbf{31.0} & 43.6 & 31.2 & 36.6 & 32.2 & 27.0 & 25.8 & 25.0 & 31.6 \\
Qwen2.5-VL 32B         & 27.6 & 30.2 & 32.8 & 30.2 & 23.6 & 21.8 & 28.8 & 31.2 & 28.3 \\
Qwen3-VL 32B           & \underline{30.0} & \underline{64.4} & \textbf{56.2} & 35.2 & \underline{33.6} & 19.4 & 28.6 & \underline{32.4} & \textbf{37.5} \\
InternVL3.5 38B        & 26.0 & 44.4 & 34.6 & \textbf{39.0} & \textbf{41.0} & 26.2 & 30.0 & \textbf{32.8} & \underline{34.3} \\
LLaVA-OV 72B           & 21.8 & 26.0 & 31.2 & 30.0 & 22.6 & 18.6 & 27.4 & 23.6 & 25.2 \\
Qwen2.5-VL 72B         & 27.4 & \textbf{66.8} & 31.8 & 31.6 & 33.0 & 24.4 & \textbf{30.4} & 25.8 & 33.9 \\
    \bottomrule
    \end{tabular}%
  }
\end{table*}

\subsection{Main Results}
\label{sec:main_results}

Table~\ref{tab:main_results} reports zero-shot accuracy across the eight SYNCR tasks. We evaluate the open-weight models on all 500 examples per task, and the proprietary models on the first 100 examples per task.
Open-weight results with 95\% confidence intervals are reported in Table~\ref{tab:full_results} in App.~\ref{app:evaluation_details}. Since SYNCR uses a four-way multiple-choice format, random chance is 25\% for all tasks except \taskNumerical, which has five options (20\%), making near-chance scores especially informative for diagnosing unresolved reasoning capabilities.

\textbf{Current MLLMs remain far below human performance.}
The strongest proprietary model, GPT-6 Astra, reaches 64.5\% average accuracy, followed by Gemini-3.1-Pro at 46.0\%, compared to the 89.5\% human baseline. Among open-weight models, Qwen3-VL 32B achieves the best overall performance with 37.5\% average accuracy, approaching the weaker closed-weight baselines but still leaving a substantial gap to human performance. These results indicate that strong general-purpose video understanding does not directly translate to robust cross-video synthesis.

\textbf{Chronological ordering is tractable, but cross-view synchronization is harder.}
The highest task-level accuracies appear in Temporal Alignment, especially on \taskOrder. GPT-6 Astra reaches 100.0\% and Gemini-3.1-Pro 96.0\%, while the best open-weight models, Qwen2.5-VL 72B and Qwen3-VL 32B, reach 66.8\% and 64.4\%, respectively. However, \taskSync~is consistently harder: the best open-weight result is only 31.0\% from InternVL3.5 14B, close to chance, and among proprietary models only GPT-6 Astra (88.0\%) and Gemini-3.1-Pro (60.0\%) clearly exceed chance. This contrast suggests that models can often recover chronological order from visible event continuity, but still struggle with precise temporal offset estimation across different camera viewpoints.

\textbf{Aggregate scores obscure task-level specialization.}
Although Qwen3-VL 32B is the strongest open-weight model overall, the best model varies substantially by task. Qwen3-VL 32B leads open-weight models only on \taskReID, InternVL3.5 38B leads on \taskMeasure, \taskNumerical, and \taskRoute, Qwen2.5-VL 72B leads on \taskOrder~and \taskCount, InternVL3.5 14B leads on \taskSync, and InternVL3.5 8B leads on \taskKinematic. This variation suggests that aggregate accuracy hides important differences in model behavior: different architectures appear to favor different forms of cross-video reasoning, and no single open-weight model dominates across all diagnostic skills.

\textbf{Fine-grained physical and global spatial reasoning remain major bottlenecks.}
The most persistent bottleneck is micro-physical reasoning, especially \taskKinematic. The best open-weight result is only 29.8\%, achieved by InternVL3.5 8B, while no proprietary model exceeds 20.0\%. Because these scores are close to the chance level of a four-way multiple-choice task, we interpret them as evidence that current models have not yet acquired reliable continuous motion comparison. The fact that the much smaller InternVL3.5 2B nearly matches this score (29.6\%) also suggests that small differences on \taskKinematic~ should be interpreted cautiously rather than as a clear scaling effect. Holistic Synthesis is similarly difficult: on \taskCount, the best open-weight model, Qwen2.5-VL 72B, reaches only 30.4\% and GPT-6 Astra 48.0\%, while on \taskRoute, the best open-weight model, InternVL3.5 38B, reaches 32.8\% and GPT-6 Astra 60.0\%. This indicates that current models struggle both to aggregate repeated object evidence and to construct a coherent global spatial map from fragmented observations.

\begin{figure}[tb]
    \centering
    \includegraphics[width=0.85\linewidth]{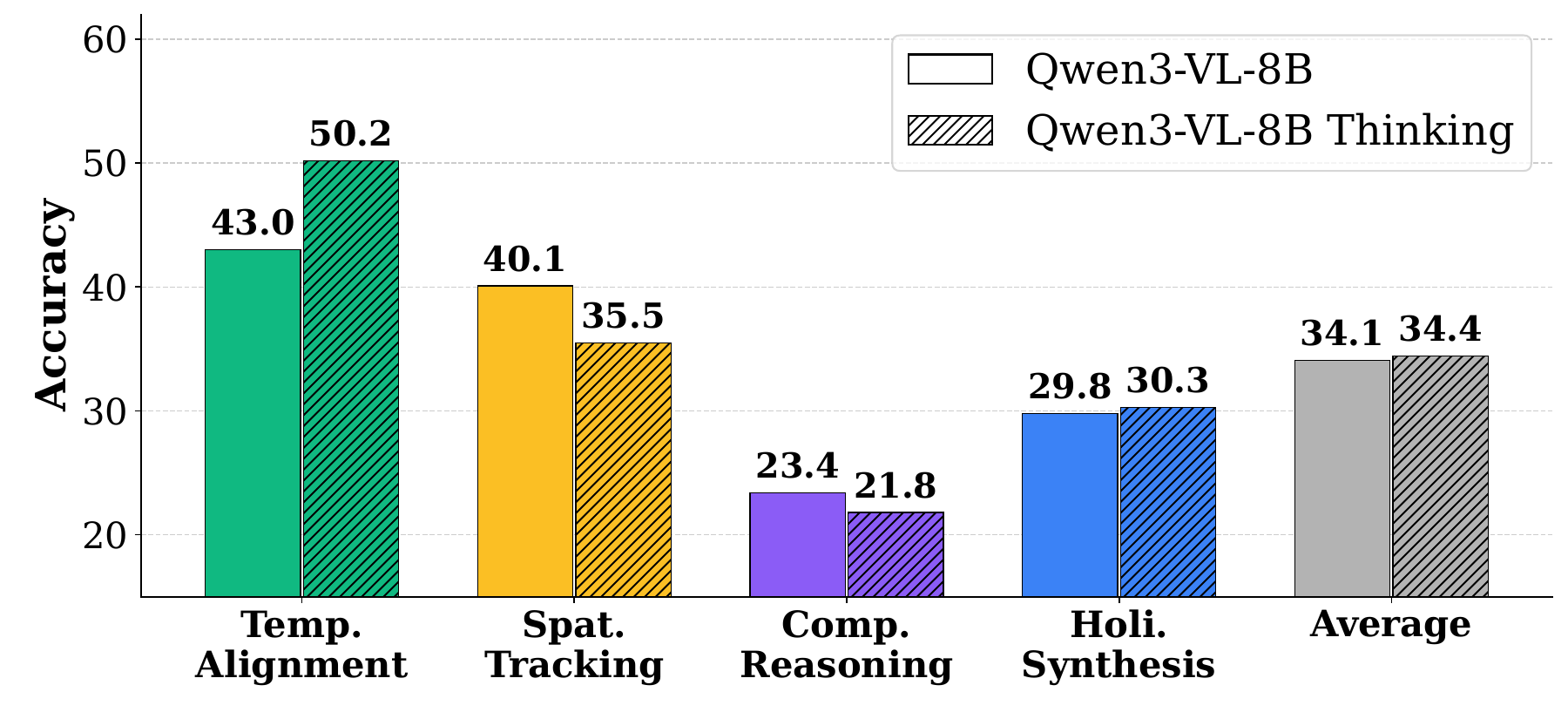}
    \caption{\textbf{Effect of reasoning-specialized post-training.}
    Qwen3-VL-8B-Thinking improves Temporal Alignment and average accuracy over Qwen3-VL-8B-Instruct, but does not improve Comparative Reasoning or Spatial Tracking, and Holistic Synthesis is nearly unchanged.}
    \label{fig:qwen_thinking_checkpoint}
\end{figure}

\textbf{Reasoning-specialized post-training helps temporal alignment but not all reasoning skills.}
We also compare Qwen3-VL-8B-Instruct with Qwen3-VL-8B-Thinking, a separately released checkpoint specialized for explicit reasoning. As shown in Fig.~\ref{fig:qwen_thinking_checkpoint}, the Thinking checkpoint improves Temporal Alignment from 43.0\% to 50.2\%, driven by \taskOrder~(56.4\% to 76.6\%), while \taskSync~decreases (29.6\% to 23.8\%). These gains do not transfer uniformly across SYNCR: Spatial Tracking decreases from 40.1\% to 35.5\% and Comparative Reasoning from 23.4\% to 21.8\%, while Holistic Synthesis remains nearly unchanged. This suggests that reasoning-specialized post-training can help with chronological alignment, but does not reliably address the fine-grained physical and multi-stream synthesis bottlenecks exposed by SYNCR.

\subsection{Scaling Trends and Task-Level Bottlenecks}
\label{sec:scaling-law}

\begin{figure}[tb]
    \centering
    \begin{subfigure}[b]{0.48\textwidth}
        \centering
        \includegraphics[width=\linewidth]{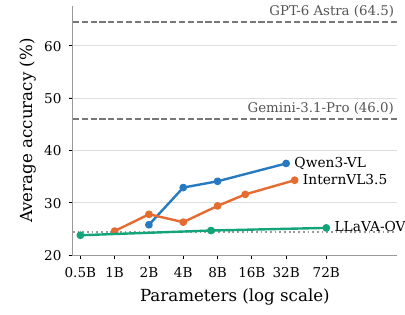}
        \caption{Average accuracy by model size}
        \label{fig:avg_scaling}
    \end{subfigure}
    \hfill
    \begin{subfigure}[b]{0.48\textwidth}
        \centering
        \includegraphics[width=\linewidth]{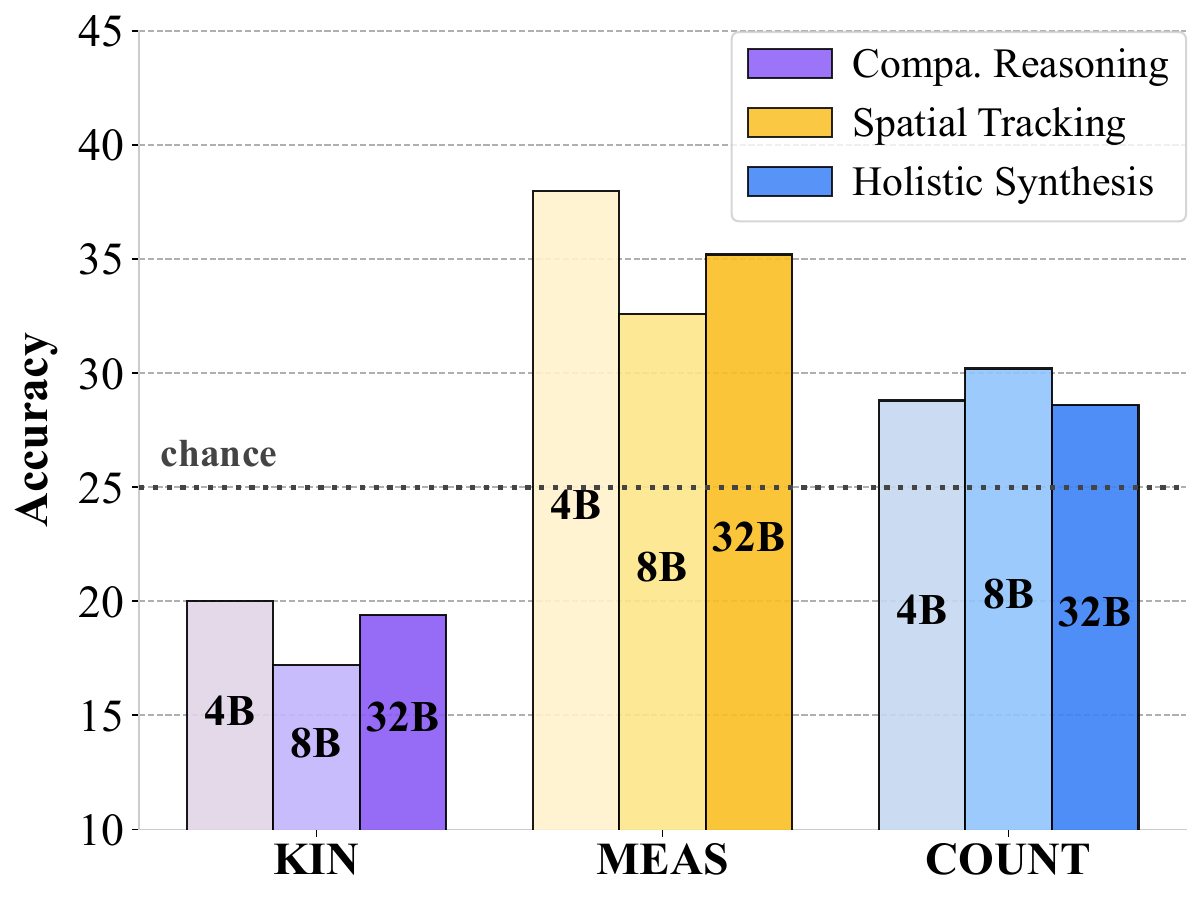}
        \caption{Tasks without consistent scaling gains}
        \label{fig:non_scaling_tasks}
    \end{subfigure}
    \caption{\textbf{Scaling behavior on SYNCR.}
    (a) Average SYNCR accuracy generally increases with model size, especially for Qwen3-VL, but gains vary across open-weight model families. Dashed lines show closed-weight baselines.
    (b) Qwen3-VL at 4B, 8B, and 32B on the three tasks that resist the family's positive average scaling trend. \taskKinematic\ stays below the 25\% chance line (dotted) at every size, while \taskMeasure\ and \taskCount\ are non-monotonic, revealing persistent bottlenecks in physical and spatial-temporal reasoning that parameter count does not address.}
    \label{fig:scaling}
\end{figure}

\textbf{Scale improves average performance, but gains are family-dependent.}
Fig.~\ref{fig:avg_scaling} visualizes average SYNCR accuracy for open-weight model families as a function of parameter scale, with closed-weight models shown as horizontal reference lines. Qwen3-VL exhibits the clearest scaling trend, improving from 25.8\% at 2B to 37.5\% at 32B.
In contrast, InternVL3.5 shows weaker and less monotonic scaling
LLaVA-OneVision shows no clear scaling, remaining between 23.8\% and 25.2\% from 0.5B to 72B, close to chance.
These trends indicate that parameter count helps on average, but scale alone does not determine cross-video reasoning performance.

\textbf{Task-level bottlenecks persist despite scaling.} Fig.~\ref{fig:non_scaling_tasks} illustrates task-level scaling behavior within the Qwen3-VL model family. Despite a positive average scaling trend, several challenging tasks exhibit clear performance plateaus. In particular, Kinematic Comparison remains close to chance across all evaluated model sizes. Furthermore, Spatial Measurement and Object Counting demonstrate limited or non-monotonic gains, indicating that parameter scaling alone does not reliably resolve bottlenecks in fine-grained physical tracking, spatial-temporal grounding, or global scene aggregation.

\subsection{Sim-to-Real Correlation Analysis}
\label{sec:correlation_analysis}

A key question for synthetic benchmarks is whether performance in simulation reflects model behavior on real-world video understanding tasks. To examine this, we compare average SYNCR accuracy with the Temporal Reasoning and Comparison tasks of MVU-Eval~\citep{peng2025mvu}. We evaluate the 17 standard open-weight checkpoints on both benchmarks with the same evaluation harness and compute Spearman rank correlations. We treat these correlations as evidence of model-level trend alignment rather than definitive evidence of transfer.

\begin{table}[t]
  \centering
  \caption{\textbf{Sim-to-real correlation.} Accuracy (\%) of the 17 standard open-weight checkpoints on SYNCR (eight-task average) and on MVU-Eval Temporal Reasoning (TR) and Comparison (Comp.), with Spearman rank correlations to the SYNCR average.}
  \label{tab:correlations}
  \adjustbox{max width=0.8\linewidth}{
    \begin{tabular}{lrrrr}
    \toprule
    \textbf{Checkpoint} & \textbf{Params (B)} & \textbf{SYNCR avg} & \textbf{MVU-Eval TR} & \textbf{MVU-Eval Comp.} \\
    \midrule
    LLaVA-OV 0.5B & 0.5 & 23.8 & 14.0 & 9.6 \\
    InternVL3.5 1B & 1 & 24.6 & 30.0 & 1.5 \\
    InternVL3.5 2B & 2 & 27.8 & 39.0 & 51.8 \\
    Qwen3-VL 2B & 2 & 25.8 & 42.5 & 56.3 \\
    InternVL3.5 4B & 4 & 26.3 & 46.0 & 60.7 \\
    Qwen3-VL 4B & 4 & 32.9 & 58.5 & 65.9 \\
    Qwen2.5-VL 7B & 7 & 26.7 & 55.0 & 61.5 \\
    LLaVA-OV 7B & 7 & 24.7 & 36.0 & 48.1 \\
    InternVL3 8B & 8 & 27.0 & 56.0 & 61.5 \\
    InternVL3.5 8B & 8 & 29.4 & 56.0 & 64.4 \\
    Qwen3-VL 8B & 8 & 34.1 & 68.0 & 68.9 \\
    InternVL3.5 14B & 14 & 31.6 & 52.5 & 67.4 \\
    Qwen2.5-VL 32B & 32 & 28.3 & 64.0 & 65.9 \\
    Qwen3-VL 32B & 32 & 37.5 & 70.5 & 71.1 \\
    InternVL3.5 38B & 38 & 34.3 & 55.0 & 65.2 \\
    Qwen2.5-VL 72B & 72 & 33.9 & 70.5 & 71.1 \\
    LLaVA-OV 72B & 72 & 25.2 & 36.0 & 23.7 \\
    \midrule
    \multicolumn{3}{l}{\textit{Spearman $\rho$ with SYNCR average}} & 0.855 & 0.917 \\
    \multicolumn{3}{l}{\textit{\quad partialling out $\log_{10}$ params}} & 0.779 & 0.870 \\
    \bottomrule
    \end{tabular}}
\end{table}

\textbf{SYNCR shows aligned model-level trends.}
As shown in Table~\ref{tab:correlations}, average SYNCR accuracy correlates strongly with both MVU-Eval tasks (Spearman $\rho=0.86$ and $0.92$, both $p<0.001$), and the correlations remain $0.78$ and $0.87$ after accounting for model size. These trends suggest that controlled SYNCR tasks capture variation across models that is also reflected in real-world multi-video evaluation.

\textbf{Correlation does not isolate cross-video reasoning.}
A general video-understanding factor could also produce these correlations, so they do not establish that SYNCR isolates cross-video reasoning. We view this agreement as evidence that SYNCR does not rank models differently from real footage, rather than as conclusive evidence of shared mechanisms.

\section{Conclusion}
In this work, we introduced SYNCR, a synthetic diagnostic framework for evaluating cross-video reasoning in MLLMs. By leveraging CLEVRER, Kubric, and Habitat, SYNCR provides programmatically verified grounding for temporal, spatial, physical, and topological variables across multiple video streams, enabling controlled evaluation across its four reasoning pillars: \textit{Temporal Alignment}, \textit{Spatial Tracking}, \textit{Comparative Reasoning}, and \textit{Holistic Synthesis}. Our zero-shot evaluation shows that current MLLMs remain far from robust cross-video reasoning: the best model reaches only 64.5\% average accuracy, compared with an 89.5\% human baseline. While models perform relatively well on chronological ordering, they struggle with precise cross-view synchronization, continuous kinematic comparison, 3D spatial measurement, and global route planning. We further find that scaling improves average performance but does not resolve the hardest physical and spatial bottlenecks, and that reasoning-specialized post-training mainly benefits chronological ordering. Finally, our sim-to-real analysis suggests that SYNCR tracks model-level trends on a real-world multi-video benchmark. Together, these results position SYNCR as a controlled testbed for diagnosing the cross-video reasoning failures that must be addressed to build more physically grounded and robust multimodal systems.

\paragraph{Limitations.}
While SYNCR provides a rigorous, mathematically verifiable environment for multi-video evaluation, our framework comes with limitations.
First, although our simulation engines (CLEVRER, Kubric, and Habitat) generate clean, programmatic physics and rendering, they are inherently limited in visual diversity and photorealism, particularly in the case of CLEVRER and Kubric. However, we frame SYNCR as a foundational 'unit test': if a model cannot perform spatial-temporal reasoning on clear, simulated objects, it will inevitably fail on unconstrained, noisy real-world footage.
Second, our evaluation methodology currently relies on multiple-choice Question Answering (QA). While this eliminates human annotation noise and ensures objective measurement, it bypasses the evaluation of open-ended conversational generation.
Third, SYNCR focuses strictly on visual, temporal, and spatial reasoning, excluding audio streams. Audio is often a critical modality for cross-modal synchronization in real-world scenarios, such as security feeds or dashcams.
Finally, while our analysis reveals critical scaling plateaus on physical reasoning tasks, our sample size of evaluated open-weight models is inherently bounded by the recent emergence of true multi-video architectures.

\newpage
\clearpage
{
\small
\bibliographystyle{ieeenat_fullname}
\bibliography{literature}
}
\newpage
\clearpage

\appendix

\begin{figure}[b]
    \centering
    \includegraphics[width=0.7\linewidth,trim={0cm 4.5cm 0cm 0cm}]{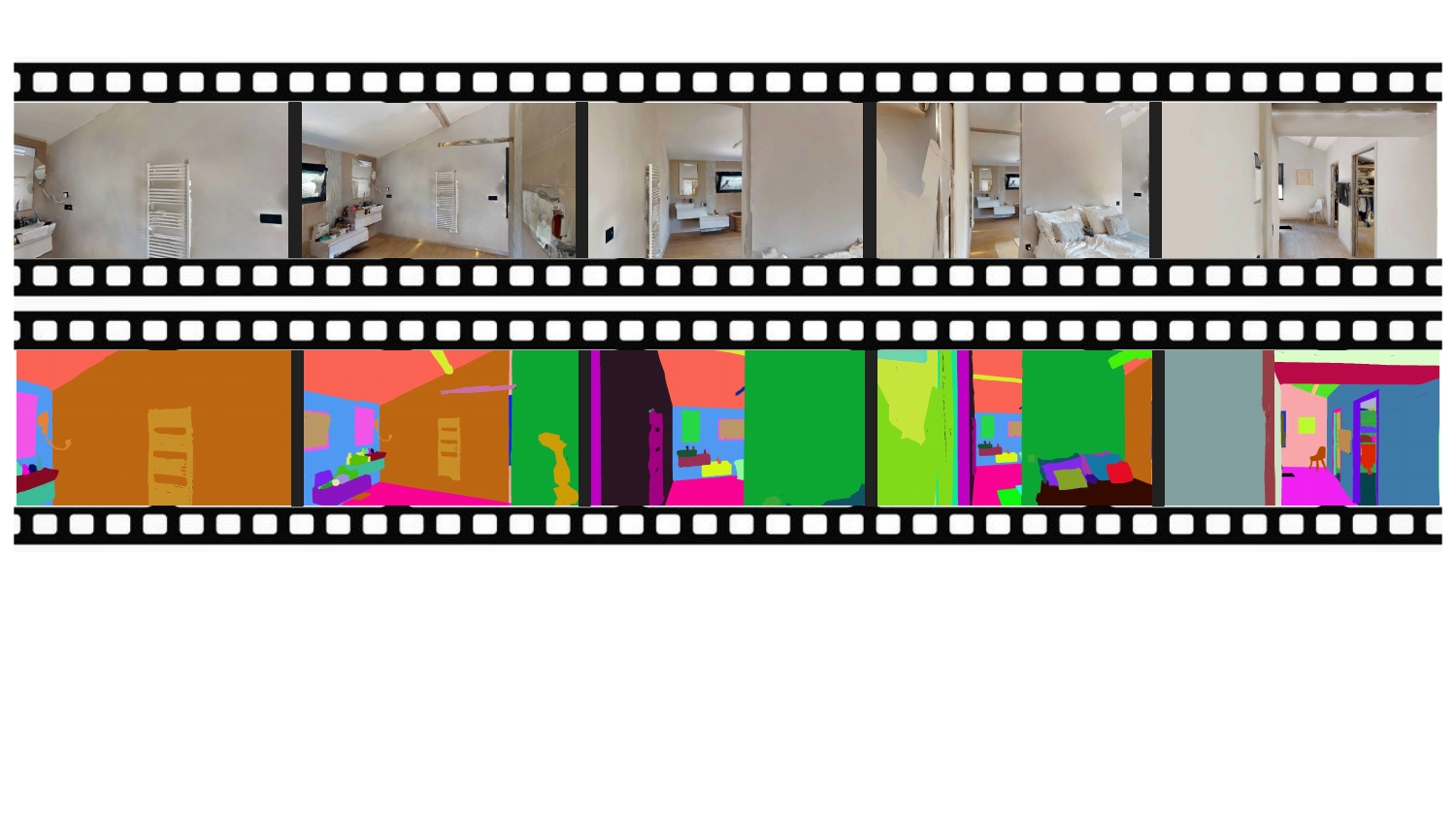}
   \caption{\textbf{Deterministic Ground Truth Generation in Habitat.}
Habitat provides synchronized RGB, depth, and semantic observations for each rendered camera pose. SYNCR uses these simulator-derived annotations to construct pixel-level object visibility, semantic instance labels, and navigation-based spatial structure for Habitat-based tasks.
}
    \label{fig:habitat-semantic}
\end{figure}

\section{Dataset Generation Details}
\label{app:dataset_details}
This section provides additional implementation details for the SYNCR data generation pipeline. We describe how each simulation environment is used to construct task-specific video inputs, extract programmatically verified ground truth, and apply filtering rules to reduce ambiguity. These details complement Sec.~\ref{sec:data} and are intended to support reproducibility of the benchmark construction process.

\subsection{Habitat Generation Details}

\textbf{Scene filtering and semantic annotations}
For each HM3D scene, we load the textured mesh, semantic annotation file, and navigation mesh. The semantic annotation file is parsed into a mapping from instance identifiers to normalized object names, while the navigation mesh is used for all pathfinding and geodesic-distance computations. As illustrated in Fig.~\ref{fig:habitat-semantic}, Habitat provides synchronized RGB, depth, and semantic observations at each camera pose, allowing us to derive task labels from simulator state rather than manual annotation. During rendering, we save the RGB stream as an MP4 video and store the corresponding semantic frame tensor separately. These semantic tensors are used only for ground-truth construction and filtering, not as model input. Scenes are skipped if the required semantic annotation or navigation files are unavailable.

\textbf{Trajectory generation}
For each sampled start--target pair, Habitat's pathfinder returns a sparse shortest path over the navigation mesh. We remove duplicate waypoints, fit a spline through the remaining path points, and snap each interpolated point back to the navigation mesh before rendering. Camera yaw is computed using a fixed look-ahead window along the trajectory to reduce frame-to-frame jitter. After the walking segment, the agent remains at the final location and performs a panoramic sweep, providing additional visual context around the endpoint.

\paragraph{Object Re-identification filtering and distractors.}
For Object Re-identification, we first identify semantic instance IDs that appear across multiple Habitat camera trajectories.
For each candidate object, we compute frame-level visibility from the semantic frame tensors and group visible frames into contiguous temporal ranges.
A candidate is retained only if it is visible for at least a minimum continuous range length and has sufficient total visibility, and only if it survives in at least two views.
The correct answer is the first contiguous timestamp window in which the same semantic instance appears in the target video.
Distractor options use the answer windows of other items with similar reference windows and lie within the 20-second video, reducing cues from option position and timing.
The corresponding prompt template is shown in Fig.~\ref{fig:prompt_habitat_reid}.

\begin{figure*}[h]
\begin{spatialbox}[Habitat Object Re-identification Template]
You are provided with two videos recorded at {fps} FPS. 
In Video 1, the '{object_name}' is clearly visible from {reference_time_window}. 
Select the correct timestamp window for the FIRST appearance of the exact same '{object_name}' in Video 2.

Options:
A) {candidate_time_window_1}
B) {candidate_time_window_2}
C) {candidate_time_window_3}
D) {candidate_time_window_4}
\end{spatialbox}
\vspace{-2mm}
\caption{\textbf{Habitat Object Re-identification prompt template.} The model is given two Habitat videos and must identify the first timestamp window in Video 2 where the same semantic instance appears.}
\label{fig:prompt_habitat_reid}
\end{figure*}

\paragraph{Object Counting filtering and balancing.}
For Object Counting, each question names two or three object categories; we enumerate semantic instance IDs for each category and compute the total count from the scene-level semantic metadata.
A question is generated only if every instance of the queried categories is visibly grounded in at least one of the input videos.
Specifically, for each target instance, at least one frame in at least one video must contain that instance with pixel occupancy of at least 5\% of the frame.
We discard cases with only one available video because the task is intended to evaluate aggregation across multiple observations, and we discard categories with more than 10 instances to avoid overly crowded counting questions.
To reduce small-count priors, we retain totals of at least $k{+}3$ for $k$ categories and balance answers across totals.
Answer choices are four distinct values between $k$ and the true count plus four.
The corresponding prompt template is shown in Fig.~\ref{fig:prompt_habitat_count}.

\begin{figure*}[h]
\begin{holisticbox}[Habitat Object Counting Template]
Based on the combined exploration in all {num_videos} videos, 
how many distinct instances of '{object_name_1}' and '{object_name_2}'
[and '{object_name_3}'] are present in this scene in total?

Options:
A) {count_option_1}
B) {count_option_2}
C) {count_option_3}
D) {count_option_4}
\end{holisticbox}
\vspace{-2mm}
\caption{\textbf{Habitat Object Counting prompt template.} The model must aggregate observations across multiple Habitat videos and count unique semantic instances without double-counting repeated views of the same object.}
\label{fig:prompt_habitat_count}
\end{figure*}
\paragraph{Route Planning path construction and distractors.}
For Route Planning, we read the route metadata associated with each generated route and reconstruct the global path by concatenating per-camera route chunks in order, removing duplicated boundary nodes between consecutive chunks.
Routes with fewer than three nodes are discarded because they do not require reasoning over an intermediate region.
The generator supports multiple path-length modes; in the default medium setting, questions are generated from subpaths whose lengths are close to the full reconstructed route length, encouraging integration across several video segments.
For each question, the correct answer is the true subpath between the selected start and end nodes.
Distractor paths use rooms observed in the input videos and match the answer hop-count distribution, with answers split evenly between two and three or more hops.
Distractors exclude direct start-to-goal paths and paths revisiting either endpoint, reducing obvious shortest-path cues.
The corresponding prompt template is shown in Fig.~\ref{fig:prompt_habitat_route}.

\begin{figure*}[h]
\begin{holisticbox}[Habitat Route Planning Template]
Based on the spatial layout shown across the videos, what is the shortest possible 
path to travel from Region 1 <{start_region_description}> to Region 2 <{end_region_description}>?

Options:
A) {candidate_path_1}
B) {candidate_path_2}
C) {candidate_path_3}
D) {candidate_path_4}
\end{holisticbox}
\vspace{-2mm}
\caption{\textbf{Habitat Route Planning prompt template.} The model must infer the shortest navigable path between two regions by integrating partial spatial observations across multiple Habitat videos.}
\label{fig:prompt_habitat_route}
\end{figure*}

\subsection{Kubric Generation Details}
To complement the overview in Section 4.2, we detail the specific simulation parameters, asset distributions, and rendering orchestration used to construct our multi-camera Kubric scenes.

\textbf{Simulation Environment and Asset Sampling.}
To ensure visual variety, the dataset is evenly split across two shape vocabularies: 1,000 scenes utilize the standard CLEVR shapes (cube, cylinder, sphere), and the remaining 1,000 scenes employ an expanded KuBasic asset suite comprising 11 distinct geometric primitives (including complex shapes like tori, gears, and teapots). To guarantee the sustained dynamic motion required for our temporal and spatial tasks, we enforce a completely frictionless floor environment with a restitution of 0.5. Objects are spawned with initial planar velocities sampled uniformly from -4.0 to 4.0 units per second along the X and Y axes, with zero initial vertical velocity to prevent immediate out-of-bounds bouncing.

\textbf{Camera Rig Configuration.}
The synchronized three-camera rig is positioned to maximize spatial coverage while offering distinct, partially overlapping viewpoints. The cameras operate with a 35mm focal length and are parameterized as follows:
\begin{itemize}
\item \textbf{Camera 0 (Right):} Positioned at base coordinates (7.48, -6.50, 5.34) and angled to look at (4, 0, 0).
\item \textbf{Camera 1 (Left):} Positioned on the opposite side of the scene at (-7.48, 6.50, 5.34) and angled to look at (-4, 0, 0).
\item \textbf{Camera 2 (Front-Center):} Positioned lower and further back at (0.0, -13.5, 5.0), directly observing the origin (0, 0, 0).
\end{itemize}

To prevent models from memorizing static camera extrinsics, a small uniform random noise vector is added to the base coordinates of each camera per scene. Finally, during the post-processing phase, pixel-level visibility matrices, 2D segmentation masks, and bounding boxes are strictly recomputed and isolated per camera view, as these visual features are inherently view-dependent despite the shared underlying 3D physics.

\paragraph{Task implementation details.}
For Kubric-based tasks, we use per-camera metadata to align object identities across views using shared asset identifiers.
Object names are generated from metadata descriptors including size, color, material, and shape.
For Multi-Angle Synchronization, we sample three independent 3-second crops from a synchronized 5-second, 12-FPS multi-camera recording and randomly assign camera views to Video 1--3.
The correct offsets are the crop-start differences of Video 2 and Video 3 relative to Video 1; distractor offsets use independent repetitions of the crop sampling, matching the answer distribution without systematic sign-flip or swap relations.
The corresponding prompt template is shown in Fig.~\ref{fig:prompt_kubric_sync}.
For Spatial Measurement, we anchor each question at the first frame where an event object exits a camera view.
Scenes with duplicate object names are discarded, target objects must be visible in all input views at the event frame, and candidate answers must be visible in at least one view.
The correct answer is the visible object with the minimum simulator-derived 3D Euclidean distance to the target object at the event frame.
Distractors match the distributions of size, shape, material, and color relations to the target; all distractors exist in the scene, and no visible distractor is closer than the answer.
The corresponding prompt template is shown in Fig.~\ref{fig:prompt_kubric_spatial_measurement}.

\begin{figure*}[h]
\begin{temporalbox}[Kubric Multi-Angle Synchronization Template]
You are provided with three videos (Video 1, Video 2, and Video 3) recorded at {fps} FPS. 
Each video is exactly {crop_duration}.0 seconds long and shows the same physical event 
captured simultaneously from three different camera angles.

Identify the exact temporal offset of the latter two videos. At what timestamp in Video 1's timeline 
does the very first frame of Video 2 and Video 3 occur? 
(Note: A negative timestamp means the video started before Video 1).

Options:
A) {offset_option_1}
B) {offset_option_2}
C) {offset_option_3}
D) {offset_option_4}
\end{temporalbox}
\vspace{-2mm}
\caption{\textbf{Kubric Multi-Angle Synchronization prompt template.} The model is given three cropped videos of the same physical event from different camera angles and must infer the temporal offsets of Video 2 and Video 3 relative to Video 1.}
\label{fig:prompt_kubric_sync}
\end{figure*}
\begin{figure*}[h]
\begin{spatialbox}[Kubric Spatial Measurement Template]
Observe the videos closely. At the exact moment the {event_object} completely exits the frame 
in Video {video_number}, which object is physically closest to the {target_object} in the 3D space?

Options:
A) {candidate_object_1}
B) {candidate_object_2}
C) {candidate_object_3}
D) {candidate_object_4}
\end{spatialbox}
\vspace{-2mm}
\caption{\textbf{Kubric Spatial Measurement prompt template.} The model is given two camera views and must identify which candidate object is closest to a target object in simulator-derived 3D space at a visually anchored event time.}
\label{fig:prompt_kubric_spatial_measurement}
\end{figure*}

\newpage
\subsection{CLEVRER Generation Details}
\label{app:clevrer_generation}

CLEVRER provides frame-level annotations for object attributes, object trajectories, velocities, visibility, and collision events.
For each split, we load the corresponding annotation directory and video directory, and associate each annotation file with its rendered video.
We use the original CLEVRER videos as visual inputs and construct cross-video question-answer pairs from the simulator annotations.
The annotations are used only for ground-truth construction and filtering, not as model input.
\paragraph{Task implementation details.}
For CLEVRER-based tasks, we use the original videos and simulator annotations containing object properties, frame-level trajectories, visibility flags, velocities, and collision events.
Annotations are used only for ground-truth construction and filtering, while models receive only the rendered videos and multiple-choice questions.
For Sequential Ordering, we split each 128-frame, 25-FPS video into \(k=4\) equal-length temporal segments separated by random gaps of 0.3--0.6 seconds, starting at a random offset, so that no two segments share a boundary frame.
The clips are shuffled, every ordering is equally frequent as an answer, and distractors are sampled in proportion to answer frequencies.
The corresponding prompt template is shown in Fig.~\ref{fig:prompt_clevrer_order}.
For Kinematic Comparison, we compute each visible object's instantaneous speed as the Euclidean norm of its 3D velocity vector and retain only pairs where the cross-video maximum velocities, and the within-video top-two velocities, differ by at least \(\tau=0.25\).
The answer choices are the top two fastest objects from each video, described by video index, color, material, and shape.
The corresponding prompt template is shown in Fig.~\ref{fig:prompt_clevrer_kinematic}.
For Numerical Comparison, we sample \(k\) distinct videos, count their annotated collision events, and ask for the video with the largest count and its difference from the second-largest count.
Tie cases use the answer format ``most and second most have the same number of collisions -- 0'', and non-tie distractors combine wrong video identities, nearby incorrect count differences, and a tie option.
Each item has five options, and items are balanced across ties and differences of one, two, and three or more, so that ties make up one fifth of the items.
The corresponding prompt template is shown in Fig.~\ref{fig:prompt_clevrer_numerical}.
\begin{figure*}[h]
\begin{temporalbox}[CLEVRER Sequential Ordering Template]
These {num_segments} videos are segments of a single continuous event, 
but they are shown out of order. What is the correct chronological order of these video segments?

Options:
A) {candidate_order_1}
B) {candidate_order_2}
C) {candidate_order_3}
D) {candidate_order_4}
\end{temporalbox}
\vspace{-2mm}
\caption{\textbf{CLEVRER Sequential Ordering prompt template.} The model is given shuffled temporal segments from one continuous CLEVRER event and must recover their original chronological order.}
\label{fig:prompt_clevrer_order}
\end{figure*}

\begin{figure*}[h]
\begin{comparativebox}[CLEVRER Kinematic Comparison Template]
Given two videos of CLEVRER physical interactions, identify which object has the highest peak velocity.

Options:
A) {candidate_video_object_1}
B) {candidate_video_object_2}
C) {candidate_video_object_3}
D) {candidate_video_object_4}
\end{comparativebox}
\vspace{-2mm}
\caption{\textbf{CLEVRER Kinematic Comparison prompt template.} The model must compare object motion across two CLEVRER videos and select the object with the highest simulator-derived peak velocity.}
\label{fig:prompt_clevrer_kinematic}
\end{figure*}
\begin{figure*}[h]
\begin{comparativebox}[CLEVRER Numerical Comparison Template]
Compare the total number of collisions across the {num_videos} videos. 
Which video has the most collisions, and what is the numerical difference 
in collisions between this video and the video with the second most collisions?

Select the correct option below.

Options:
A) {collision_option_1}
B) {collision_option_2}
C) {collision_option_3}
D) {collision_option_4}
E) {collision_option_5}
\end{comparativebox}
\vspace{-2mm}
\caption{\textbf{CLEVRER Numerical Comparison prompt template.} The model must compare collision counts across CLEVRER videos and identify both the video with the most collisions and the count difference relative to the second-highest video.}
\label{fig:prompt_clevrer_numerical}
\end{figure*}

\newpage
\section{Experimental Details}
\label{app:evaluation_details}
\textbf{Evaluation Protocol.}
All models are evaluated in a zero-shot multiple-choice setting.
Each benchmark example is stored as a JSONL record containing a natural-language question, four answer options (five for \taskNumerical), the ground-truth answer string, and a list of input videos.
For every example, we construct the evaluation prompt shown in Fig.~\ref{fig:prompt_eval}.
The answer options are presented in the same order stored in the benchmark record, and the ground-truth answer is stored using the same formatted option string, e.g., ``A) ...''.

The input videos are passed to the model in the order specified by the example's \texttt{video\_list}.
Before each video, we insert a text marker of the form ``Video 1:'', ``Video 2:'', etc., so that textual references in the question are aligned with the visual inputs.
If a video record contains crop metadata, such as \texttt{start\_sec} and \texttt{end\_sec}, only the corresponding temporal segment is provided to the model; otherwise, the full video is used. Timestamps within each crop are rebased to start at zero, so they do not reveal the crop's position in the source video.
This allows the same evaluation interface to support both full-video tasks and tasks such as CLEVRER Sequential Ordering or Kubric Multi-Angle Synchronization, where each input corresponds to a temporal crop of a longer source video.
We use deterministic decoding for all models.
For open-weight models, generation is run with sampling disabled and temperature set to 0 whenever the model interface exposes a temperature parameter.
We score the model's final answer with a shared parser: when an answer marker is present, its final stated option is compared with the ground truth; otherwise, the parser accepts an explicit option commitment, a single option-lettered answer line, or a response matching one option's text. Unparsed responses are counted as incorrect.

\begin{figure*}[h]
\begin{promptbox}[Evaluation Prompt]
{question}

Options:
A) {option 1}
B) {option 2}
C) {option 3}
D) {option 4}
[E) {option 5}]

Format Instructions:
Provide your final answer in its own line and using exactly this format: 
Answer: <option_alphabet>) <full_option_text>
\end{promptbox}
\vspace{-2mm}
\caption{\textbf{Prompt for evaluation.} We use the same multiple-choice evaluation prompt for all SYNCR tasks.}
\label{fig:prompt_eval}
\end{figure*}

\textbf{Statistical Significance.} To rigorously assess the reliability of our findings and contextualize the performance gaps between models, we report 95\% confidence intervals for the open-weight accuracy scores. Because our evaluation protocol employs deterministic decoding (temperature set to 0), response variance cannot be measured across multiple generation passes. Instead, we estimate the uncertainty due to item sampling using 95\% Wilson score intervals~\citep{wilson1927probable}, reported as a margin of error ($\pm$) in our final results. At chance, the interval half-width is approximately 3.8 points for 500 items.
\begin{table}[htbp]
  \centering
  \color{black}
  \captionsetup{font+=appendixblackfont,labelfont+=appendixblackfont}
  \caption{\textbf{Full-set results with confidence intervals.} \newtext{Accuracy (\%) on all 500 items per task for each standard open-weight checkpoint, with 95\% Wilson confidence intervals.}}
  \label{tab:full_results}
  \adjustbox{max width=\textwidth}{
    \begin{tabular}{L{26mm}*{8}{C{17mm}}C{7mm}}
    \toprule
    \multirow{2}{*}{\textbf{Model}} & \multicolumn{2}{c}{\textbf{\tacS}} & \multicolumn{2}{c}{\textbf{\stcS}} & \multicolumn{2}{c}{\textbf{\corcS}} & \multicolumn{2}{c}{\textbf{\hscS}} & \multirow{2}{*}{\textbf{Avg}} \\
    \cmidrule(lr){2-3} \cmidrule(lr){4-5} \cmidrule(lr){6-7} \cmidrule(lr){8-9}
    & \textsc{Sync} & \textsc{Order} & \textsc{ReID} & \textsc{Meas}  & \textsc{Num} & \textsc{Kin} & \textsc{Count} & \textsc{Route} & \\
    \midrule
    \multicolumn{10}{l}{\textbf{Model Size $\leq$ 4B}} \\
    \midrule
    LLaVA-OV 0.5B & 25.8 $\pm$ 3.8 & 24.4 $\pm$ 3.8 & 26.4 $\pm$ 3.9 & 23.0 $\pm$ 3.7 & 16.0 $\pm$ 3.2 & 24.0 $\pm$ 3.7 & 25.4 $\pm$ 3.8 & 25.4 $\pm$ 3.8 & 23.8 \\
    InternVL3.5 1B & 24.4 $\pm$ 3.8 & 25.8 $\pm$ 3.8 & 24.6 $\pm$ 3.8 & 24.4 $\pm$ 3.8 & 20.0 $\pm$ 3.5 & 27.6 $\pm$ 3.9 & 22.8 $\pm$ 3.7 & 27.4 $\pm$ 3.9 & 24.6 \\
    InternVL3.5 2B & 26.8 $\pm$ 3.9 & 25.2 $\pm$ 3.8 & 26.0 $\pm$ 3.8 & 31.2 $\pm$ 4.0 & 26.8 $\pm$ 3.9 & \underline{29.6 $\pm$ 4.0} & 28.4 $\pm$ 3.9 & 28.0 $\pm$ 3.9 & 27.8 \\
    Qwen3-VL 2B & 24.2 $\pm$ 3.7 & 28.4 $\pm$ 3.9 & 33.6 $\pm$ 4.1 & 29.4 $\pm$ 4.0 & 21.4 $\pm$ 3.6 & 20.2 $\pm$ 3.5 & 25.6 $\pm$ 3.8 & 23.2 $\pm$ 3.7 & 25.8 \\
    InternVL3.5 4B & 26.4 $\pm$ 3.9 & 30.6 $\pm$ 4.0 & 26.2 $\pm$ 3.8 & 24.8 $\pm$ 3.8 & 29.2 $\pm$ 4.0 & 24.2 $\pm$ 3.7 & 23.6 $\pm$ 3.7 & 25.6 $\pm$ 3.8 & 26.3 \\
    Qwen3-VL 4B & 26.2 $\pm$ 3.8 & 49.4 $\pm$ 4.4 & 46.4 $\pm$ 4.4 & \underline{38.0 $\pm$ 4.2} & 26.6 $\pm$ 3.9 & 20.0 $\pm$ 3.5 & 28.8 $\pm$ 4.0 & 28.0 $\pm$ 3.9 & 32.9 \\
    \midrule
    \multicolumn{10}{l}{\textbf{Model Size 4B $<$ size $\leq$ 8B}} \\
    \midrule
    LLaVA-OV 7B & 27.2 $\pm$ 3.9 & 21.6 $\pm$ 3.6 & 26.6 $\pm$ 3.9 & 27.6 $\pm$ 3.9 & 20.8 $\pm$ 3.6 & 24.2 $\pm$ 3.7 & 26.0 $\pm$ 3.8 & 23.6 $\pm$ 3.7 & 24.7 \\
    Qwen2.5-VL 7B & 25.0 $\pm$ 3.8 & 36.8 $\pm$ 4.2 & 25.8 $\pm$ 3.8 & 29.8 $\pm$ 4.0 & 22.4 $\pm$ 3.6 & 21.2 $\pm$ 3.6 & 27.2 $\pm$ 3.9 & 25.6 $\pm$ 3.8 & 26.7 \\
    InternVL3 8B & 26.0 $\pm$ 3.8 & 35.4 $\pm$ 4.2 & 25.0 $\pm$ 3.8 & 29.6 $\pm$ 4.0 & 26.6 $\pm$ 3.9 & 26.0 $\pm$ 3.8 & 27.2 $\pm$ 3.9 & 20.0 $\pm$ 3.5 & 27.0 \\
    InternVL3.5 8B & 26.4 $\pm$ 3.9 & 39.4 $\pm$ 4.3 & 26.0 $\pm$ 3.8 & 31.4 $\pm$ 4.1 & 31.6 $\pm$ 4.1 & \textbf{29.8 $\pm$ 4.0} & 26.4 $\pm$ 3.9 & 24.0 $\pm$ 3.7 & 29.4 \\
    Qwen3-VL 8B & 29.6 $\pm$ 4.0 & 56.4 $\pm$ 4.3 & \underline{47.6 $\pm$ 4.4} & 32.6 $\pm$ 4.1 & 29.6 $\pm$ 4.0 & 17.2 $\pm$ 3.3 & \underline{30.2 $\pm$ 4.0} & 29.4 $\pm$ 4.0 & 34.1 \\
    \midrule
    \multicolumn{10}{l}{\textbf{Model Size $>$ 8B}} \\
    \midrule
    InternVL3.5 14B & \textbf{31.0 $\pm$ 4.0} & 43.6 $\pm$ 4.3 & 31.2 $\pm$ 4.0 & 36.6 $\pm$ 4.2 & 32.2 $\pm$ 4.1 & 27.0 $\pm$ 3.9 & 25.8 $\pm$ 3.8 & 25.0 $\pm$ 3.8 & 31.6 \\
    Qwen2.5-VL 32B & 27.6 $\pm$ 3.9 & 30.2 $\pm$ 4.0 & 32.8 $\pm$ 4.1 & 30.2 $\pm$ 4.0 & 23.6 $\pm$ 3.7 & 21.8 $\pm$ 3.6 & 28.8 $\pm$ 4.0 & 31.2 $\pm$ 4.0 & 28.3 \\
    Qwen3-VL 32B & \underline{30.0 $\pm$ 4.0} & \underline{64.4 $\pm$ 4.2} & \textbf{56.2 $\pm$ 4.3} & 35.2 $\pm$ 4.2 & \underline{33.6 $\pm$ 4.1} & 19.4 $\pm$ 3.5 & 28.6 $\pm$ 3.9 & \underline{32.4 $\pm$ 4.1} & 37.5 \\
    InternVL3.5 38B & 26.0 $\pm$ 3.8 & 44.4 $\pm$ 4.3 & 34.6 $\pm$ 4.2 & \textbf{39.0 $\pm$ 4.3} & \textbf{41.0 $\pm$ 4.3} & 26.2 $\pm$ 3.8 & 30.0 $\pm$ 4.0 & \textbf{32.8 $\pm$ 4.1} & \underline{34.3} \\
    LLaVA-OV 72B & 21.8 $\pm$ 3.6 & 26.0 $\pm$ 3.8 & 31.2 $\pm$ 4.0 & 30.0 $\pm$ 4.0 & 22.6 $\pm$ 3.7 & 18.6 $\pm$ 3.4 & 27.4 $\pm$ 3.9 & 23.6 $\pm$ 3.7 & 25.2 \\
    Qwen2.5-VL 72B & 27.4 $\pm$ 3.9 & \textbf{66.8 $\pm$ 4.1} & 31.8 $\pm$ 4.1 & 31.6 $\pm$ 4.1 & 33.0 $\pm$ 4.1 & 24.4 $\pm$ 3.8 & \textbf{30.4 $\pm$ 4.0} & 25.8 $\pm$ 3.8 & 33.9 \\
    \bottomrule
    \end{tabular}%
  }
\end{table}

\paragraph{Compute and Hardware.} All zero-shot evaluations were conducted using deterministic decoding (temperature set to 0.0). Because the SYNCR benchmark requires processing multiple independent video streams simultaneously (ranging from 2 to 4 videos per question), memory limits and inference latency are primary considerations. All open-weight model evaluations were executed on single compute nodes with NVIDIA A100 (80GB) or H100 (80GB) GPUs, using tensor parallelism for models that do not fit on one GPU. Moreover, Table.~\ref{tab:runtime} details the average inference latency for a single question-answer sample for the timed open-weight models across all eight SYNCR tasks.
\begin{table}[htbp]
    \centering
\caption{\textbf{Average Per-Sample Inference Latency on SYNCR.}
Inference time (in seconds) across four cognitive pillars and eight tasks. We report the open-weight MLLMs grouped by parameter scale. All evaluations were conducted using a dual NVIDIA A100 (80GB) setup using Tensor Parallelism where necessary. Closed-weight proprietary systems are excluded as they were evaluated via API. \newtext{Qwen2.5-VL 7B and 32B and InternVL3 8B were not timed. The largest Qwen3-VL entries (8B on \textsc{Order}; 32B on \textsc{Sync} and \textsc{Kin}) most likely reflect occasional long generated reasoning rather than input size, since Qwen3-VL often writes out its reasoning before answering. Qwen3-VL-8B-Thinking, whose completions had a median of about 5,200 tokens in a small pilot, took roughly three minutes per \textsc{Sync} item, against 2.6 seconds for the Instruct model.}}
  \label{tab:runtime}
  \adjustbox{max width=\textwidth}{
    \begin{tabular}{L{40mm}*{8}{C{10mm}}C{6mm}}
    \toprule
    \multirow{2}{*}{\textbf{Model}} & \multicolumn{2}{c}{\textbf{\tacS}} & \multicolumn{2}{c}{\textbf{\stcS}} & \multicolumn{2}{c}{\textbf{\corcS}} & \multicolumn{2}{c}{\textbf{\hscS}} & \multirow{2}{*}{\textbf{Avg}} \\
    \cmidrule(lr){2-3} \cmidrule(lr){4-5} \cmidrule(lr){6-7} \cmidrule(lr){8-9}
    & \textsc{Sync} & \textsc{Order} & \textsc{ReID} & \textsc{Meas}  & \textsc{Num} & \textsc{Kin} & \textsc{Count} & \textsc{Route} & \\
    \midrule
    \multicolumn{10}{l}{\textbf{Model Size $\leq$ 4B}} \\
    \midrule
LLaVA-OV 0.5B \href{https://huggingface.co/llava-hf/llava-onevision-qwen2-0.5b-ov-hf}{(Link)}          & 3.4 & 1.3 & 1.0 & 0.4 & 4.0 & 0.7 & 1.1 & 4.0 & 2.0 \\
InternVL3.5 1B \href{https://huggingface.co/OpenGVLab/InternVL3_5-1B}{(Link)}         & 10.9 & 16.8 & 5.6 & 1.4 & 3.6 & 2.9 & 7.8 & 9.1 & 7.3 \\
InternVL3.5 2B \href{https://huggingface.co/OpenGVLab/InternVL3_5-2B}{(Link)}         & 13.5 & 25.1 & 5.3 & 1.5 & 3.1 & 2.9 & 8.2 & 9.3 & 8.6 \\
Qwen3-VL 2B \href{https://huggingface.co/Qwen/Qwen3-VL-2B-Instruct}{(Link)}            & 1.5 & 1.7 & 1.5 & 0.6 & 0.9 & 1.0 & 2.0 & 3.3 & 1.6 \\
InternVL3.5 4B \href{https://huggingface.co/OpenGVLab/InternVL3_5-4B}{(Link)}         & 8.3 & 16.5 & 6.4 & 2.1 & 3.7 & 3.7 & 9.3 & 11.3 & 7.7 \\
Qwen3-VL 4B \href{https://huggingface.co/Qwen/Qwen3-VL-4B-Instruct}{(Link)}            & 2.4 & 2.2 & 1.8 & 1.0 & 1.1 & 1.2 & 2.2 & 3.6 & 1.9 \\
    \midrule
    \multicolumn{10}{l}{\textbf{4B $<$ Model Size $\leq$ 8B}} \\
    \midrule
LLaVA-OV 7B \href{https://huggingface.co/llava-hf/llava-onevision-qwen2-7b-ov-hf}{(Link)}            & 1.9 & 2.2 & 1.6 & 1.0 & 0.9 & 1.1 & 1.7 & 1.7 & 1.5 \\
InternVL3.5 8B \href{https://huggingface.co/OpenGVLab/InternVL3_5-8B}{(Link)}         & 9.4 & 18.3 & 6.8 & 2.6 & 4.4 & 4.1 & 10.2 & 12.0 & 8.5 \\
Qwen3-VL 8B \href{https://huggingface.co/Qwen/Qwen3-VL-8B-Instruct}{(Link)}            & 2.6 & 16.1 & 2.2 & 1.1 & 8.2 & 1.4 & 2.5 & 4.1 & 4.8 \\
    \midrule
    \multicolumn{10}{l}{\textbf{Model Size $>$ 8B}} \\
    \midrule
InternVL3.5 14B \href{https://huggingface.co/OpenGVLab/InternVL3_5-14B}{(Link)}        & 13.6 & 24.7 & 8.5 & 3.6 & 5.2 & 5.3 & 12.4 & 14.5 & 11.0 \\
Qwen3-VL 32B \href{https://huggingface.co/Qwen/Qwen3-VL-32B-Instruct}{(Link)}           & 99.0 & 4.7 & 3.7 & 2.1 & 3.3 & 38.4 & 4.2 & 6.9 & 20.3 \\
InternVL3.5 38B \href{https://huggingface.co/OpenGVLab/InternVL3_5-38B}{(Link)}        & 39.1 & 53.7 & 18.4 & 10.7 & 13.5 & 13.1 & 27.6 & 31.1 & 25.9 \\
LLaVA-OV 72B \href{https://huggingface.co/llava-hf/llava-onevision-qwen2-72b-ov-hf}{(Link)}           & 9.8 & 11.5 & 6.4 & 5.2 & 6.4 & 5.7 & 8.0 & 11.9 & 8.1 \\
Qwen2.5-VL 72B \href{https://huggingface.co/Qwen/Qwen2.5-VL-72B-Instruct}{(Link)}         & 42.2 & 34.9 & 6.2 & 4.3 & 34.8 & 40.5 & 7.3 & 11.8 & 22.8 \\
    \bottomrule
    \end{tabular}%
  }
\end{table}
\newpage
\section{Human Baseline Details}
\label{app:human_baseline}

To contextualize model performance, we conduct a small human reference evaluation on a balanced subset of SYNCR.
As described in Sec.~\ref{sec:experimental_setup}, we use the first 100 question-answer pairs from each of the eight SYNCR tasks, the same items given to the proprietary models, yielding 800 examples in total.
The sample is balanced across tasks so that each diagnostic pillar contributes equally to the final estimate.

Four graduate-student volunteers participated in the human evaluation.
All participants had prior experience with machine learning or computer vision research, but they were not given access to simulator annotations, ground-truth labels, or dataset-generation metadata. Each participant was shown the same inputs given to the models; only the rendered videos and the corresponding multiple-choice questions. Each of the four participants answered all 800 examples independently.
For each example, we compute the final human prediction by majority vote over the four responses.
If all annotators do not agree, the option selected by the largest number of annotators is used.
In the rare case of a two-way tie, we mark the example as incorrect for the aggregate human baseline to avoid inflating human performance.

The human evaluation is used only to estimate a reference accuracy for the benchmark.
It does not involve collecting personal, sensitive, demographic, or behavioral information from participants.
The videos are simulator-rendered: Habitat traverses HM3D scans of real indoor spaces, whereas Kubric and CLEVRER depict generated object scenes.
No participant data beyond the selected answer choices is used in the analysis.

\end{document}